\documentclass[final]{cvpr}

\usepackage{times}
\usepackage{epsfig}
\usepackage{graphicx}
\usepackage{amsmath}
\usepackage{amssymb}

\renewcommand{\vec}[1]{\mathbf{#1}}

\usepackage{tabularx}
\usepackage{multirow}
\usepackage{amsmath}
\usepackage{amssymb}
\usepackage{graphicx}
\usepackage{xcolor}
\usepackage{colortbl}
\usepackage{stmaryrd}
\usepackage{amsthm}
\usepackage{bm}
     
\newtheorem{thm}{Theorem}[section]

\newtheorem{prop}[thm]{Proposition}

\newtheorem{mydef}[thm]{Definition}

\renewcommand{\vec}[1]{\mathbf{#1}}

\newcommand{\Rd}{\mathbb{R}^d}
\newcommand{\R}{\mathbb{R}}

\newcommand{\X}{\mathcal{X}}
\newcommand{\Sn}{S^n}

\newcommand{\U}{\mathcal{U}}

\newcommand{\argmaxUnique}[2]{\underset{#1}{\operatorname{argmax }}\, #2}

\newcommand{\Hi}{\mathcal{H}}

\newcommand{\la}{\left\langle}
\newcommand{\ra}{\right\rangle}

\usepackage[pagebackref=true,breaklinks=true,colorlinks,bookmarks=false]{hyperref}

\begin{document}

\title{Kernelized Classification in Deep Networks}

\author{Sadeep Jayasumana ~~~~~~ Srikumar Ramalingam ~~~~~~ Sanjiv Kumar\\ 
Google Research, New York \\
\tt\small{\{sadeep,rsrikumar,sanjivk\}@google.com}
}

\maketitle

\begin{abstract}
We propose a kernelized classification layer for deep networks. Although conventional deep networks introduce an abundance of nonlinearity for representation (feature) learning, they almost universally use a linear classifier on the learned feature vectors.  We advocate a nonlinear classification layer by using the kernel trick on the softmax cross-entropy loss function during training and the scorer function during testing. However, the choice of the kernel remains a challenge. To tackle this, we theoretically show the possibility of optimizing over all possible positive definite kernels applicable to our problem setting. This theory is then used to device a new kernelized classification layer that learns the optimal kernel function for a given problem automatically within the deep network itself. We show the usefulness of the proposed nonlinear classification layer on several datasets and tasks.
\end{abstract}

\section{Introduction}
Deep learning is built
upon the premise that useful representations of the inputs can be automatically learned
from data~\cite{bengio2013}. For example, in the image classification setting, a rich representation learning network consisting of building blocks such as convolution and max-pooling is first used to obtain a vector representation of the input image. This representation is commonly referred to as the \emph{feature vector} of the input image. 

The image is then classified into the correct class within the last layer of the network using a fully-connected layer operating on the feature vector~\cite{vggnet, resnet}. This last classification layer represents a linear classifier in the space of the learned feature vectors. Therefore, to perform well on the classification task, the classes have to be linearly separable in the space of feature vectors. While this is a standard assumption in many tasks, it is conceivable that using a nonlinear classifier on the learned feature vectors would give additional benefits, especially when the backbone feature extractor does not have enough capacity to learn fully linearly separable features.

Kernel methods is a different branch of machine learning that has been successful in the pre-deep-learning era, particularly with the popularity of the Support Vector
Machines (SVM) algorithm~\cite{Cortes1995, Scholkopf2002book}. Conventionally, kernel methods have been used for learning with hand-crafted feature vectors, such as histogram-of-oriented-gradients (HOG)~\cite{Dalal05Hog} and bag-of-visual-words~\cite{vedaldi2009} for image classification. The key idea in kernel methods 
is the following: instead of running a linear classifier on the feature vectors, they are first
mapped to a higher-dimensional Reproducing Kernel Hilbert Space (RKHS) using a positive definite kernel function. A linear classifier is then run on this high-dimensional RKHS.
Since the dimensionality of the feature vectors is dramatically increased via this mapping, a linear classifier in the RKHS corresponds to a powerful nonlinear classifier in the original feature vector space. 
Thanks to the kernel trick, we never have to
explicitly calculate the high-dimensional vectors in the RKHS, which
will be computationally expensive to compute and store.

Although kernel methods yield excellent results in shallow machine learning in general, the choice of the kernel function is
often problematic. There is a collection of well-known kernels such as the linear kernel, polynomial
kernels, and the Gaussian RBF kernel~\cite{Shawe-Taylor2004book, Scholkopf2002book}. The Gaussian RBF kernel is a common
choice; however, tuning its bandwidth parameter is non-trivial and different values for this parameter can give vastly
different results. Furthermore, the Gaussian RBF kernel is only one member of the family of positive definite kernels - a different kernel might well be the best fit for a given problem.

In this work, we join the concepts of automatic representation learning and
nonlinear, kernel-based classification. To this end, we use a nonlinear, kernelized classification layer in deep networks. Importantly, we address the problem of choosing an appropriate kernel for the task by theoretically showing that it is possible to sweep over the entire class of positive definite kernels applicable to our problem setting to automatically find the optimal kernel. With this result, we device a kernel learning method that learns the optimal kernel from data within the deep learning framework itself. The full network, which consists of a conventional representation learner and our kernelized classification layer powered with automatic kernel learning, can be trained end-to-end using the usual backpropagation algorithm. 

In addition to the theoretical formulation and the algorithm for learning the optimal kernel, we also want to study the role of the proposed kernelized classifier layer in improving generalization, finding compact student networks with distillation, and training accurate models with less data. To this end, we show the benefits of this framework in image classification, transfer learning, distillation, and active learning settings on a number of datasets.

\section{Related Work}
One of the earliest attempts to connect neural networks and kernel methods was the sigmoid kernel~\cite{vapnik_book}, which became popular in SVMs due to the early success of the neural networks. This kernel was inspired by the sigmoid activation used in the early generations of neural networks. More recently, the authors of~\cite{NIPS2009_3628} proposed a family of kernel functions that mimic the computation in multilayer neural nets, and showed their usage in multilayer kernel networks.

There have been a few methods that focus on extending the linear convolution in CNNs to a nonlinear operation. Convolutional kernel networks~\cite{MairalKHS14} provide a kernel approximation scheme to interpret convolutions. Volterra series approximations were used in \cite{Zoumpourlis2017} to extend convolutions to a nonlinear operation by introducing quadratic terms. The authors of \cite{wang2019kervolutional} proposed a kernelized version of the convolution operation and demonstrated that it can learn more complicated features than the usual convolution. Our work differs from theirs in a number of ways: Some kernels used in their work, such as the $L^p$-norm kernels, are not positive definite~\cite{Berg84} and therefore do not represent a valid RKHS mapping~\cite{Aronszajn1950}. In contrast, we strictly work with positive definite kernels, which represent valid mappings to an RKHS. Furthermore, learning of hyperparameters of pre-defined kernels advocated in their work is principally different from the kernel learning method presented in this paper -- we theoretically show that our method optimizes over the space of \emph{all} radial positive definite kernels on the unit sphere to find the best kernel, instead of limiting the optimization to the hyperparameters of pre-defined kernels. 

An RBF kernel layer was introduced in \cite{chen2019deep} to produce a feature space from pointcloud input. In contrast to our work, the RBF kernel layer is used on the input data and not in the classification layer.

There have been several interesting explorations of loss functions other than the usual softmax crossentropy loss in CNN settings, specially in the open-set classification setting~\cite{Wen_IJCV_2019, Cevikalp_2021_PAMI, Deng_CVPR_2019, Wang_CVPR_2018}. An example is the Center Loss~\cite{Wen_IJCV_2019}, which encourages low intra-class variance in the feature vectors. Other methods such as~\cite{Deng_CVPR_2019} achieve higher performance by leaving additional margins in the softmax loss. Our work differs from these since we work in the closed-set classification setting and employ an automatically learned kernel to obtain nonlinear classification boundaries.

Second order pooling methods~\cite{Lin_ICCV_2015, Li_ICCV_2017, Wang_PAMI_2020} are another way of performing nonlinear classification in the feature vector space. In has been shown that second order pooling is equivalent to using a second-degree polynomial kernel function~\cite{Gao_CVPR_2016, Cai_ICCV_2017}. The authors of \cite{Cui2017} extended second-order pooling to higher orders while learning the coefficients of higher order interactions. Their method, however, requires explicit calculation of feature maps, which they tackle using Fast Fourier Transforms. In a work parallel to ours~\cite{Mahmoudi_ICIP_2020}, the kernel trick is used in the dense layer along with the polynomial kernel. Our work differs from the above works in that we never calculate explicit feature maps and we theoretically show that our method learns over the space of all possible positive definite kernels on the hyper-sphere, which includes polynomial kernels.

Prior to the dominance of deep learning methods, picking the right kernel for a given problem has been studied extensively in works such as ~\cite{Howley2005, Ali2006, Jayasumana_CVPR2014, Gonen11multiplekernel}. In particular, Multiple Kernel Learning (MKL) approaches~\cite{Gonen11multiplekernel, Varma07learningthe} were popular in conjunction with SVM. Unfortunately, these methods scale poorly with the size of the training dataset. In this work, we automatically learn the kernel within a deep network. This not only allows automatic representation learning, but also scales well for large training sets.

Kernels have also been considered for deep learning to reduce the memory footprint of CNNs. This was accomplished by achieving an end-to-end training of a Fastfood kernel layer~\cite{Yang2014}, which uses approximations of kernel functions using Fastfood transforms~\cite{Le2013}. Other related methods involving both kernels and deep learning include 
scalable kernel methods~\cite{dai2015scalable}, deep SimNets~\cite{CohenSS15}, and deep kernel learning~\cite{WilsonHSX15}.
\section{Nonlinear Softmax Loss}
\label{sec:nl_softmax}
Let us consider a multi-class classification problem with a training set
$\left\{(x_i, y_i)\right\}_{i=1}^N$, where $x_i \in \X$ for each $i$, $y_i \in [L] \stackrel{.}{=} \{1, 2, \dots, L \}$ for each
$i$, $\X$ is a nonempty set, $L$ is the number of labels, and $N$ is the number of training examples. For instance, each training data point $(x_i, y_i)$ can be an image with its class label.

A deep neural network that solves this task has two components: a
\emph{representation learner} (also called the \emph{feature learner}) and a \emph{classifier}. In the case of image classification, the representation learner  consists of modules such as convolution layers, max-pooling layers, and fully-connected layers. The classifier is the last fully-connected layer operating on the feature vectors, which is endowed with a loss function during training.

Let $r^{(\Theta)}:
\mathcal{X} \to \R^d$ denote the representation learner, where $d$ is the dimensionality of the learned feature vectors and $\Theta$ represents all the (learnable) parameters in this part of the network. The classifier is charaterized by a function $g^{(\Omega)}: \Rd \to [L]$, where $\Omega$ denotes all the parameters in the last layer of the network. Usually, $\Omega$ consists of weight vectors $\vec{w}_1, \vec{w}_2, \dots, \vec{w}_L$ with each $\vec{w}_j \in \R^d$, and bias terms $b_1, b_2, \dots, b_L$ with each $b_j \in \R$. The function $g^{(\Omega)}$ then takes the form:
\vspace{-0.15cm}
\begin{equation}
\label{eq:scorer_with_f}
    g^{(\Omega)}(\mathbf{f}) = \argmaxUnique{j}{\;\mathbf{w}_j^T\mathbf{f}},
\end{equation}
where $\vec{f} = r^{(\Theta)}(x) \in \Rd$ is the feature vector for input $x$. Note that we have dropped the additive bias term $b_j$ to keep the notation uncluttered. There is no loss of generality here since the bias term can be absorbed into $\vec{w}_j$ by appending a constant element to $\vec{f}$. During inference, the deep network's class prediction $\hat{y}^{*}$ for an input $x^{*}$ is the composite of these two functions:
\vspace{-0.15cm}
\begin{equation}
\label{eq:scorer}
    \hat{y}^{*} = \left(g^{(\Omega)} \circ r^{(\Theta)}\right)(x^{*}).
\end{equation}
Although conceptually there are two components of the deep network, their parameters $\Theta$ and $\Omega$ are learned jointly during the training of the network. The de facto standard way of training a classification network is minimizing the \emph{softmax loss} applied to the classification layer. The softmax loss is the combination of the softmax function and the cross-entropy loss. More specifically, for a single
training example $(x, y)$ with the feature vector $\vec{f} = r^{(\Theta)}(x)$, the softmax loss is calculated as,
\vspace{-0.15cm}
\begin{equation}
\label{eq:softmax}
l(y, \vec{f}) = -\log\left(\frac{\exp(\vec{w}_y^T\vec{f})}{\sum_{j=1}^L
\exp(\vec{w}_j^T\vec{f})}\right).
\end{equation}
Note that the classifier $g^{(\Omega)}$ trained is this manner is completely linear in $\R^d$, the
space of the feature vectors $\vec{f}$s, as is evident from~\eqref{eq:scorer_with_f}. 

From the classic knowledge in kernel methods, we are aware that more powerful nonlinear classifiers on
$\R^d$ can be obtained using the kernel trick. The key idea here is to first embed
the feature vectors $\vec{f}$s into a high-dimensional Reproducing Kernel Hilbert Space (RKHS) $\Hi$ and
perform classification in $\Hi$. Although the classification is linear in the high-dimensional Hilbert space $\Hi$, it is nonlinear in
the original feature vector space $\Rd$. Let $\phi:
\R^d \to \Hi$ represent this RKHS embedding. Performing classification in $\Hi$ is then equivalent to training the neural network with the following modified version
of the softmax loss:
\vspace{-0.15cm}
\begin{equation}
\label{eq:nl_softmax}
l'(y, \vec{f}) = -\log\left( \frac{ \exp\left(\la \phi(\vec{w}_y), \phi(\vec{f}) \ra_\Hi \right)} {\sum_{j=1}^L \exp\left(\la\phi(\vec{w}_j), \phi(\vec{f})\ra_\Hi \right)}\right),
\end{equation}
where $\la .,. \ra_\Hi$ denotes the inner product in the Hilbert space $\Hi$. The key difference between \eqref{eq:softmax} and \eqref{eq:nl_softmax} is that the dot products between $\vec{w}_j$s and $\vec{f}$ have been replaced with the inner products between $\phi(\vec{w}_j)$s and $\phi(\vec{f})$. The more general notion of \emph{inner product} is used instead of the \emph{dot product} because the Hilbert space $\Hi$ can be infinite dimensional.

For a network trained with this nonlinear softmax function, predictions can be obtained using a modified version of the predictor:
\vspace{-0.15cm}
\begin{equation}
\label{eq:nl_predictor}
g'^{(\Omega)}(\vec{f}) = \argmaxUnique{j}{ \la \phi(\vec{w}_j), \phi(\vec{f}) \ra_\Hi}.
\end{equation}
Note that the Hilbert space embeddings $\phi(.)$s can be very-high, even infinite, dimensional. Therefore, computing and storing them can be problematic. We can use the \emph{kernel trick} in the classic machine learning literature~\cite{Scholkopf2002book, Shawe-Taylor2004book} to overcome this problem: explicit computation of $\phi(.)$s is avoided by directly evaluating the inner product between them using a kernel function $k : \Rd \times \Rd \to \R$. That is:
\begin{equation}
\la \phi(\vec{w}), \phi(\vec{f}) \ra_\Hi = k(\vec{w}, \vec{f}).
\end{equation}
However, for a kernel function to represent a valid RKHS, it must be positive definite~\cite{Aronszajn1950, Berg84}. We discuss this notion next.

\section{Kernels on the Unit Sphere}
\label{sec:kernels_on_sphere}
It was shown in the previous section that, given a kernel function on the feature vector space, we can obtain a nonlinear classifier in the last layer of a deep network by modifying the softmax loss function during training and the predictor during inference. Only positive definite  kernels allow this trick. There are various choices for kernel functions in the classic machine learning literature. Some popular choices include the polynomial kernel, the Gaussian RBF kernel (squared exponential kernel), and the Laplacian kernel. However, in the classic kernel methods literature, there is no principled method for selecting the optimal kernel for a given problem. Furthermore, many of the kernels have hyperparameters that need to be manually tuned. The generally accepted solution to this problem in classic kernel methods is the MKL framework~\cite{Gonen11multiplekernel}, where the optimal kernel is learned as a linear combination of some pre-defined kernels. Unfortunately, like SVM, MKL methods do not scale well with the train set size.

In this section, we present some theoretical results that will pave the way to define a neural network layer that can automatically learn the optimal kernel from data. By formulating kernel learning as a neural network layer, we inherit the desirable properties of deep learning, including scalability and automatic feature learning. Importantly, we show that our method can sweep over the entire space of positive definite kernels applicable to our problem setting to find the best kernel.

We start the discussion with the following definition of positive definite  kernels~\cite{Berg84}.
\begin{mydef}\label{def:posdefKernels}
Let $\U$ be a nonempty set. A function $k: (\U \times \U) \to
\R$ is called a \textbf{positive definite  kernel} if $k(u, v) = k(v, u)$ for all $u, v \in \U$ and
\[\sum_{j=1}^N\sum_{i = 1}^{N}c_ic_jk(u_i, u_j) \ge 0,\]
for all $N \in \mathbb{N}, \{u_1, \ldots, u_N\} \subseteq \U$ and $\{c_1,\dots, c_N\} \subseteq \R$.
\end{mydef}

Properties of positive definite kernels have been studied extensively in mathematics literature~\cite{Aronszajn1950, Schoenberg1938, Berg84}. The following proposition summarizes some important closure properties of this class of functions.

\begin{prop}
\label{prop:closure}
The family of all positive definite kernels on a given nonempty set forms a convex cone that is closed under pointwise multiplication and pointwise convergence.
\end{prop}
\begin{proof}
To intuitively understand this result, it is helpful to recall that the geometry of the family of the all positive definite kernels on a given nonempty set is closely related to the geometry of the space of the $p \times p$ symmetric positive definite matrices, which forms a convex cone. The formal proof of this proposition can be found in Remark 1.11 and Theorem 1.12 of Chapter 3 of ~\cite{Berg84}.
\end{proof}

To simplify the problem setting, we assume that both the feature vectors $\vec{f}$s and the weight vectors $\vec{w}_j$s are $L^2$-normalized. Not only this simplifies the mathematics, but also it is a practice in use for stabilizing the training of neural networks~\cite{Yi:2019, Liu_CVPR_2017, Hoffer_ICLR_2018}. Due to this assumption, we are interested in positive definite kernels on the unit sphere in $\Rd$. From now on, we use $\Sn$, where $n = d - 1$, to denote this space.

We also restrict our discussion to radial kernels on $\Sn$. Radial kernels, kernels that only depend on the distance between the two input points, have the desirable property of translation invariance. Furthermore, all the commonly used kernels on $\Sn$, such as the linear kernel, the polynomial kernel, the Gaussian RBF kernel, and the Laplacian kernel are radial kernels. The following theorem, origins of which can be traced back to \cite{Schoenberg1942Shpere}, fully characterizes radial kernels on $\Sn$.

\begin{thm}
\label{thm:main}
A radial kernel $k : \Sn \times \Sn \to \R$ is positive definite for any
$n$ if and only if it admits a unique series representation of the form
\begin{align}
\label{eq:infinite_series}
k(\vec{u}, \vec{v}) = &\sum_{m=0}^\infty \alpha_m \la \vec{u}, \vec{v} \ra^m \nonumber\\
& + \;\alpha_{-1} (\llbracket \la \vec{u}, \vec{v} \ra = 1\rrbracket -
\llbracket \la \vec{u}, \vec{v} \ra = -1\rrbracket ) \nonumber\\
& + \;\alpha_{-2} \llbracket \la \vec{u}, \vec{v} \ra \in \{-1,
1\}\rrbracket,
\end{align}
where each $\alpha_m \ge 0\,$, $\sum_{m=-2}^\infty \alpha_m < \infty$, and
$\llbracket . \rrbracket$ depicts the Iversion bracket.
\end{thm}
\begin{proof}
The kernel $k_1: \Sn \times \Sn \to [-1, 1]:  k_1(\vec{u}, \vec{v}) = \la \vec{u}, \vec{v} \ra$ is positive definite on $\Sn$ for any $n$ since $\sum_j\sum_i c_i c_j \la \vec{u}_i, \vec{u}_j \ra = \| \sum_i c_i \vec{u}_i \|^2 \ge 0$. Therefore, from the closure properties in
Proposition~\ref{prop:closure}, the kernel $k_m: (\vec{u}, \vec{v}) \mapsto \la \vec{u}, \vec{v} \ra^m$ is also positive definite on $\Sn$ for
any $m \in \mathbb{N}$. Furthermore, $k_m$ is positive definite for $m = 0$ since $\sum_j\sum_i c_i c_j \la \vec{u}_i, \vec{u}_j \ra^0 = \left\| \sum_i c_i  \right\|^2 \ge 0$.

Let us now consider the following two sequences of kernels:
\begin{align*}
s_{\rm{odd}} &= k_1, \;k_3,\; \dots,\; k_{2m + 1}, \; \dots\;\; \text{and}\\
s_{\rm{even}} &= k_2, \; k_4, \; \dots, \; k_{2m}, \; \dots
\end{align*}
Since $-1 \le \la \vec{u}, \vec{v} \ra \le 1$, it is clear that $s_{\rm{odd}}$ and $s_{\rm{even}}$
converge pointwise to the following kernels, respectively.
\begin{align*}
k_{\rm{odd}}(\vec{u}, \vec{v}) &= \llbracket \la \vec{u}, \vec{v} \ra = 1\rrbracket - \llbracket\la \vec{u}, \vec{v} \ra = -1\rrbracket,\\
k_{\rm{even}}(\vec{u}, \vec{v}) &= \llbracket \la \vec{u}, \vec{v} \ra \in \{-1, 1\}\rrbracket.
\end{align*}
From the last closure
property of Proposition~\ref{prop:closure}, both $k_{\rm{odd}}$ and $k_{\rm{even}}$ are
positive definite on $S^n$. Invoking Proposition~\ref{prop:closure} again, we conclude that any finite conic combination of the kernels $k_{\rm{even}}, k_{\rm{odd}}, k_0, k_1, \dots$ is positive definite on $S^n$ for
any $n$. This completes the forward direction of the proof. 

For the proof of the converse, we refer the reader to Theorem 3.6 in
Chapter 5 of \cite{Berg84}.\end{proof}

Equipped with a complete characterization of the positive definite radial kernels on $\Sn$, we now discuss how we can combine this result with the nonlinear softmax formulation in \S~\ref{sec:nl_softmax} to automatically learn the best kernel classifier within a deep network.
\section{The Kernelized Classification Layer}
We now introduce a kernelized classification layer that acts as a drop-in replacement for the usual softmax classification layer in a deep network. This new layer classifies feature vectors in a high-dimensional RKHS while automatically choosing the optimal positive definite kernel that enables the mapping into the RKHS. As a result, we do not have to hand-pick a kernel or its hyperparameters.

\subsection{Mechanics of the Layer}
Our classification layer is parameterized by the usual weight vectors: $\vec{w}_1, \vec{w}_2, \dots, \vec{w}_L$, and some additional learnable coefficients: $\alpha_{-2}, \alpha_{-1}, \dots, \alpha_M$, where $M \in \mathbb{N}$ and each $\alpha_m \ge 0$. During training, this classifier maps feature vectors $\vec{f}$s to a high-dimensional RKHS $\Hi_{\rm opt}$, which optimally separates feature vectors belonging to different classes, and learns a linear classifier in $\Hi_{\rm opt}$. During inference, the classifier maps feature vectors of previously unseen input to the RKHS it learned during training and performs classification in that space. This is achieved by using the nonlinear softmax loss defined in \eqref{eq:nl_softmax} during training and the nonlinear predictor defined in \eqref{eq:nl_predictor} during  testing, with the inner product in $\Hi$ given by:
\begin{equation}
\label{eq:opt_kernel}
\la \phi(\vec{w}), \phi(\vec{f}) \ra_{\Hi} = \la \phi(\vec{w}), \phi(\vec{f}) \ra_{\Hi_{\rm opt}} = k_{\rm opt}(\vec{w}, \vec{f}),
\end{equation}
where $k_{\rm opt}(., .)$ is the reproducing kernel of $\Hi_{\rm opt}$. The optimal RKHS $\Hi_{\rm opt}$ for a given classification problem is learned by finding the optimal kernel $k_{\rm opt}$ during training as discussed in the following.

Theorem~\ref{thm:main} states that any positive definite radial kernel on $\Sn$ admits the series representation shown in \eqref{eq:infinite_series}. Therefore, the optimal kernel $k_{\rm opt}$ must also have such a series representation. We approximate this series with a finite summation by cutting off the terms beyond the order $M$:

\begin{align}
\label{eq:finite_series}
    k_{\rm opt}(\vec{w}, \vec{f}) \approx \sum_{m=0}^M \alpha_m  &k_m(\vec{w}, \vec{f}) + \alpha_{-1}k_{\rm odd}(\vec{w}, \vec{f}) \nonumber\\
    &+ \alpha_{-2}k_{\rm even}(\vec{w}, \vec{f}),
\end{align}
where, $k_{\rm{even}}, k_{\rm{odd}}, k_0, k_1, \dots, k_M$ have meanings defined in \S~\ref{sec:kernels_on_sphere} and $\alpha_{-2}, \alpha_{-1}, \dots, \alpha_{M} \ge 0$. Using Proposition~\ref{prop:closure} and the discussion in the proof of Theorem~\ref{thm:main}, one can easily verify that this approximation does not violate the positive definiteness of $k_{\rm opt}$.

With this, $k_{\rm opt}$ is learned automatically from data by making the coefficients $\alpha_{-2}, \alpha_{-1}, \dots, \alpha_{M}$s learnable parameters of the classification layer. Let $\bm{\alpha} = [\alpha_{-2}, \alpha_{-1}, \dots, \alpha_{M}]^T$. The gradient of the loss function with respect to $\bm{\alpha}$ can be calculated via the backpropagation algorithm using \eqref{eq:nl_softmax}, \eqref{eq:opt_kernel}, and \eqref{eq:finite_series}. Therefore, it can be optimized along with $\vec{w}_1, \vec{w}_2, \dots, \vec{w}_L$ during the gradient descent based optimization of the network. This procedure is equivalent to automatically finding the RKHS that optimally separates the feature vectors belonging to different classes.

The constraint $\alpha_{-2}, \alpha_{-1}, \dots, \alpha_{M} \ge 0$ in \eqref{eq:finite_series} can be imposed using the methods discussed in \S~\ref{sec:alpha_reg}. As for the number of kernels $M$ in the approximation, as long as it is sufficiently large, the exact value of $M$ is not critical. This is because, as discussed in the proof of Theorem~\ref{thm:main}, the higher order terms that are truncated approach either $k_{\rm odd}$ or $k_{\rm even}$, both of which are already included in the finite summation. On the other hand, if the terms beyond some order $M' < M$ are not important, the network can automatically learn to make the corresponding $\alpha$ coefficients vanish. In practice, we observed that 10 kernels work well enough and stick to this number in all our experiments. 

Importantly, the kernelized classification layer described above can pass on the gradients of the loss to its inputs: the feature vectors $\vec{f}$s. Therefore, our kernelized classification layer is fully compatible with end-to-end training and can act as a drop-in replacement for an existing softmax classification layer.

\subsection{Regularization of the Coefficients}
\label{sec:alpha_reg}
The constraint $\alpha_{-2}, \alpha_{-1}, \dots, \alpha_{M} \ge 0$ is important to preserve the positive definiteness of $k_{\rm opt}$. This can be imposed by using $\bm{\alpha} = \rm{ReLU}(\bm\alpha')$, where $\bm\alpha'$ is the learnable parameter vector. However, $\rm{ReLU}$ has no upper-bound and allowing the scale of $\bm{\alpha}$ to grow unboundedly causes issues in optimization: Assume we have an instantiation ${\bm{\alpha}_0}$ of the vector $\bm{\alpha}$. By replacing ${\bm{\alpha}_0}$ with $\lambda{\bm{\alpha}_0}$, where $\lambda > 1$, we scale all the inner product terms in \eqref{eq:nl_softmax} and \eqref{eq:nl_predictor} by the same $\lambda$. As a result, we improve the loss of the already correctly classified training examples, but without making any effective change to the predictor. 
Therefore, under this setting, once the majority of the training examples are correctly classified, the neural network can easily improve the loss just by increasing the norm of $\bm\alpha$, which is not useful. We therefore advocate an $L^2$-regularization term on $\bm{\alpha}$ when $\rm{ReLU}$ activation is used.

Alternatively, one could also use $\bm{\alpha} = \rm{sigmoid}(\bm\alpha')$ or $\bm{\alpha} = \rm{softmax}(\bm\alpha')$, both of which not only guarantee $\alpha_{-2}, \alpha_{-1}, \dots, \alpha_{M} \ge 0$, but also produce bounded $\bm{\alpha}$. Therefore, no regularization on $\bm\alpha$ is needed for these options. The $\rm{softmax}$ activation here should not be confused with the softmax loss discussed in \S~\ref{sec:nl_softmax}. The usage of the $\rm{softmax}$ activation in this context is similar to that in the self-attention literature~\cite{Vaswani_2017}, where it is used to normalize the coefficients of a linear combination. 

\subsection{Usage in the Distillation Setting}
\label{sec:distillation}
We expect the kernelized classification to be particularly useful in settings where the capacity of the feature learning or backbone network is capacity limited. This is because a capacity-limited network might not be able to learn fully linearly separable features and therefore a nonlinear classifier can be useful to augment its capabilities.

Another common method used to improve classification with capacity-limited networks is knowledge distillation~\cite{Hinton_distil}, where the logit or probability outputs of a larger \emph{teacher} network is used to train a smaller \emph{student} network. While training the student network, the loss function (partially) consists of the cross-entropy loss with the teacher network's output. More specifically, using the same notation as in \S~\ref{sec:nl_softmax}, assume that for a training example $(x, y)$, the teacher produces logits $\vec{h} = [h_1, h_2, \dots, h_L]^T$. Then, for the student network with the feature vector $\vec{f} = r^{(\Theta)}(x)$ and a usual classification layer parameterized by the weight vectors $\vec{w}_1, \vec{w}_2, \dots, \vec{w}_L$,  the cross-entropy loss is given by:
\begin{equation}
\label{eq:linear_distillation}
    l_{\rm{st}}(\vec{h}, \vec{f}) = - \sum_{j'=1}^L \tilde{h}_{j'} \log\left(\frac{\exp(\vec{w}_{j'}^T\vec{f}/T)}{\sum_{j=1}^L \exp(\vec{w}_j^T\vec{f}/T)}\right),
\end{equation}
where $\tilde{h}_{j'} = \exp(h_{j'} / T)/ {\sum_{j=1}^L \exp(h_{j}/T)}$ and $T$ is the temperature hyperparameter.

In distillation, the student network tries to imitate a teacher network, which is capable of producing more powerful feature vectors than the student. Intuitively, therefore, the student could benefit from using a powerful nonlinear classifier on the weak feature vectors it produces. With this in mind, we explore the use of kernelized classification layer in the student network. The cross-entropy loss with the teacher scores in this case is:
\begin{equation}
\label{eq:nl_distillation}
l'_{\rm{st}}(\vec{h}, \vec{f}) = - \sum_{j'=1}^L \tilde{h}_{j'} \log\left( \frac{ \exp\left(\la \phi(\vec{w}_{j'}), \phi(\vec{f}) \ra_\Hi/T \right)} {\sum_{j=1}^L \exp\left(\la\phi(\vec{w}_j), \phi(\vec{f})\ra_\Hi/T \right)}\right),
\end{equation}
where all the terms have the meanings defined earlier.

\begin{figure}[t]
\setlength{\tabcolsep}{1pt}
\centering
 \begin{tabular}{c  c} 
     \includegraphics[width=0.24\textwidth]{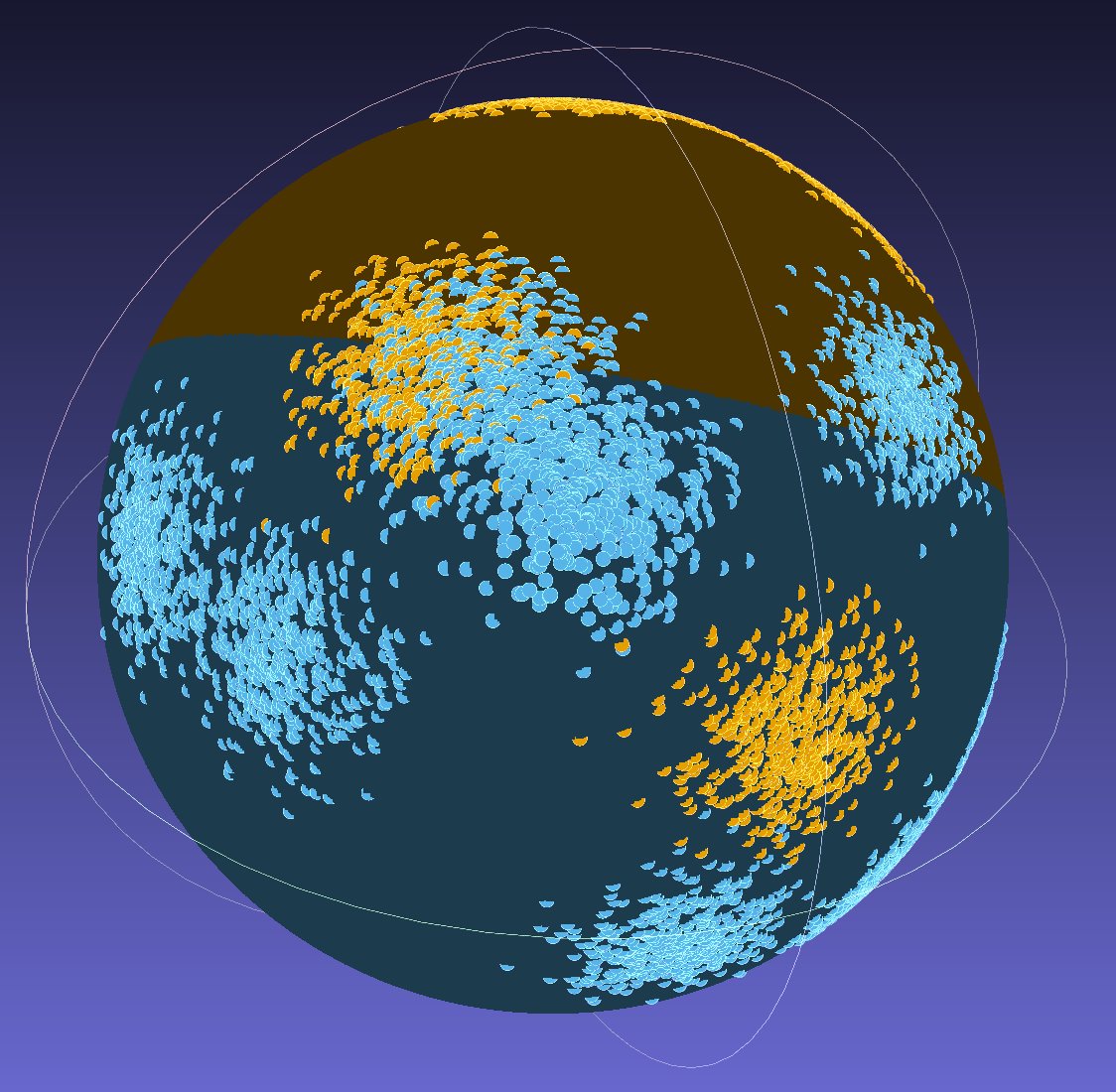} &
     \includegraphics[width=0.24\textwidth]{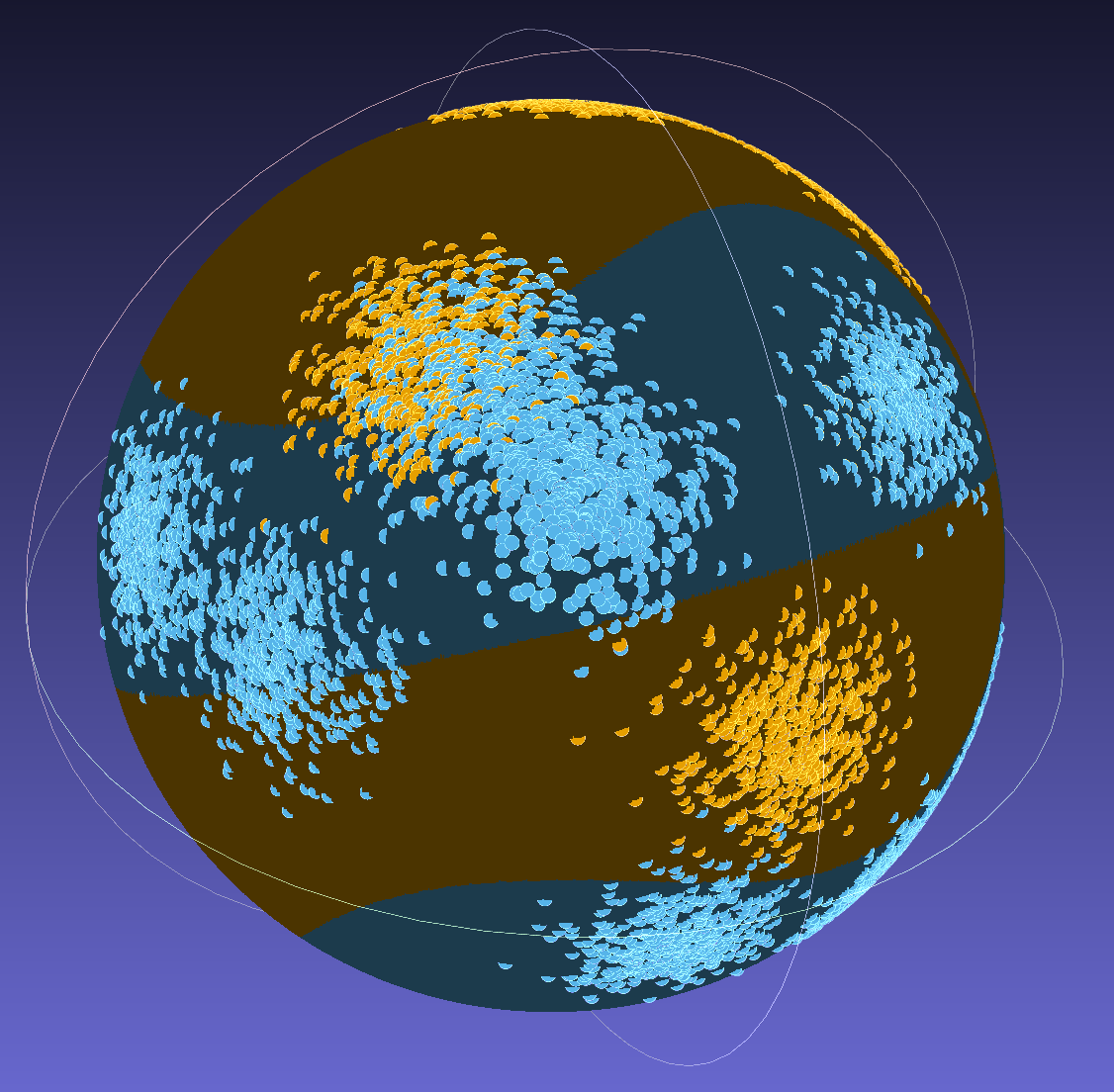} \\
     {\small (a) Softmax classifier} & {\small(b) Kernelized classifier}
 \end{tabular}
\caption{\label{fig:synthetic_data} {\bf Classification of a binary dataset on $S^2$.} {\it Regions identified by the classifiers and training data are shown in blue and orange colors. Note that the softmax classifier can only separate cap-like regions on the sphere, whereas our kernelized classifer can do more complex nonlinear classification thanks to the higher dimensional RKHS embedding of the sphere. Best viewed in color.\vspace{-0.2cm}}}
\end{figure}

\section{Experiments}
For all experiments, the main baseline is the standard softmax classifier. In image classification, we show three additional baselines based on the linear kernel~\cite{Hoffer_ICLR_2018}, second order pooling~\cite{Lin_ICCV_2015}, and kervolutional networks~\cite{wang2019kervolutional}.

To evaluate our method, we replace the classification layer with our proposed kernelized classification layer. In all experiments we use 10 kernels, meaning that the kernelized classification layer uses only 10 additional learnable parameters. As discussed in \S~\ref{sec:alpha_reg}, we use $\rm{ReLU}$ activation and a weight decay of $0.0001$ on this 10-dimensional parameter vector. This is the same amount of weight decay used in the other parts of the network. This vector is initialized with all ones.

To keep the number of learnable parameters comparable, we keep the additive bias term in the baseline classifier and omit them in the kernelized classifier. This bias term introduces a number of learnable parameters equal to the number of classes used. Therefore, in most cases, the baseline model actually uses more learnable parameters than the kernelized classifier. We also remove the $\operatorname{ReLU}$ activation on the feature vectors to utilize the full surface of $\Sn$. The same is done to the baseline model to enable a fair comparison (see \S~\ref{sec:ablation_study} for more details). More details of our experiment setup are given in the Appendix.

\subsection{Synthetic Data}
\label{sec:exp:synthetic}
We first evaluate the proposed kernelized classification layer as an isolated unit by demonstrating its capabilities to learn nonlinear patterns on $\Sn$. To this end, inspired by the blue-orange dataset in \cite{Hastie_statisticallearning}, we generated a two-class dataset on $S^2$ with a mixture of Gaussian clusters for each class. The data generation process is detailed in the Appendix.

We treat generated observations on $S^2$ as feature vectors and train the softmax classification layer (baseline) and our kernelized classification layer on them for comparison. Results on the test set are shown in Table~\ref{tbl:synthetic_data}. We also report the theoretical maximum accuracy, the accuracy of the Bayes optimal classifier. It is noteworthy that the accuracy of our kernelized classification layer significantly outperforms the baseline and gets close to the Bayes optimal performance. This can be attributed to the layer's capabilities to learn nonlinear patterns on the sphere by embedding the data into an RKHS that optimally separates different classes.

We visualize our training data, the baseline classifier's output and the kernelized classifier's output in Figure~\ref{fig:synthetic_data}. Note that the usual softmax classifier can only separate cap-like regions on $S^2$, this is a result of its being a linear classifier with respect to the feature vectors. Our kernelized classifier, on the other hand, can do a more complex nonlinear separation of the data.

\begin{table}[h]
\centering
 \begin{tabular}{|l | c |} 
 \hline
 \multicolumn{1}{|c|}{Method}& Accuracy \\
 \hline
 Softmax classifier (baseline)  & 85.51  \\ 
 Kernelized classifier (ours) & {\bf 94.20} \\
 Bayes optimal classifier & \cellcolor{gray!25}{95.06} \\
 \hline
 \end{tabular}
 \vspace{0.2cm}
 \caption{\label{tbl:synthetic_data}{\bf Results on the synthetic dataset.} {\it Note that the accuracy of the kernelized classifier is close to that of the ideal Bayes optimal classifier (theoretical maximum).}}
\end{table}

\subsection{Image Classification}
We now report results on CIFAR-10 and CIFAR-100 real world image benchmarks~\cite{Krizhevsky09Cifar}. For each dataset, we experimented with several CIFAR ResNet architectures~\cite{resnet}. We considered four different baselines: (1) SM: the standard softmax loss, (2) LIN: normalized feature vectors and weights with only the linear kernel along with a learnable coefficient. This is similar to the approach discussed in~\cite{Hoffer_ICLR_2018}, but with additional freedom to learn the weight vectors, (3) SOP: second order pooling~\cite{Lin_ICCV_2015}, which is also equivalent to~\cite{Mahmoudi_ICIP_2020} with a second degree polynomial, and (4) KER: kervolutional networks~\cite{wang2019kervolutional} with the Gaussian RBF kernel.

Accuracy figures are summarized in Tables~\ref{tbl:cifar10_icml}~and~\ref{tbl:cifar100_icml}. In all cases our method significantly outperforms the baselines. This shows the benefits of optimizing over the entire space of positive definite kernels instead of restricting ourselves to linear methods or pre-defined kernels.
\begin{table}[h!]
\centering
 \begin{tabular}{|l | c | c | c | c | c |} 
 \hline
 Backbone  &  SM & LIN & SOP & KER &  Ours\\
 \hline
 ResNet-8  & 83.73 & 82.45 & 84.03 & 85.15 & \bf{86.93} \\ 
 ResNet-14 & 89.87 & 90.16 & 90.47 & 90.41 & \bf{91.48} \\
 ResNet-20 & 91.14 & 91.01 & 91.75 & 91.34 & \bf{92.88} \\
 ResNet-32 & 92.22 & 92.21 & 92.31 & 92.41 & \bf{93.70} \\
 ResNet-44 & 92.10 & 93.10 & 92.42 & 92.87 & \bf{94.05} \\
 ResNet-56 & 93.01 & 93.13 & 93.33 & 93.09 & \bf{94.15} \\
 \hline
 \end{tabular}
 \vspace{0.2cm}
 \caption{\label{tbl:cifar10_icml}{\bf Results on the CIFAR-10 dataset.}}
\end{table}

\begin{table}[h!]
\centering
 \begin{tabular}{|l | c | c | c | c | c |} 
 \hline
 Backbone  &  SM & LIN & SOP & KER & Ours\\
 \hline
 ResNet-8  & 53.82 & 54.00 & 55.80 & 56.86 & \bf{58.27} \\ 
 ResNet-14 & 63.85 & 63.67 & 63.53 & 64.14 & \bf{66.67} \\
 ResNet-20 & 65.99 & 65.79 & 67.97 & 67.31 & \bf{69.33} \\
 ResNet-32 & 68.96 & 69.16 & 70.39 & 69.40 & \bf{71.30} \\
 ResNet-44 & 70.16 & 70.54 & 71.13 & 71.09 & \bf{73.20} \\
 ResNet-56 & 71.23 & 72.11 & 73.12 & 72.39 & \bf{74.10} \\
 \hline
 \end{tabular}
 \vspace{0.2cm}
 \caption{\label{tbl:cifar100_icml}{\bf Results on the CIFAR-100 dataset.}}
\end{table}

\subsection{Transfer Learning}
In the next set of experiments we evaluate our method in a transfer learning setting. To this end, we take a ResNet-50 network pre-trained on the Imagenet ILSVRC 2012 classification dataset~\cite{imagenet_cvpr09} and fine tune it on Oxford-IIIT Pets~\cite{Parkhi_2012} and Stanford Cars~\cite{KrauseStarkDengFei} datasets. For each dataset, we use the train/test splits provided by the standard Tensorflow Datasets implementation~\cite{tensorflow_datasets}. Results are summarized in Table~\ref{tbl:transfer_learning}. Note that the KER baseline is not possible in this setting as it involves changes to the backbone network. 
On both datasets, kernelized classification layer results in significant gains over the baselines. This is intuitive to understand since the feature vectors learned from the source task (Imagenet) might not linearly separate the new classes in the target task. We can therefore benefit from a nonlinear classifier in the transfer learning setting.
\begin{table}[h!]
\centering
 \begin{tabular}{|l | c | c | c | c |} 
 \hline
 \multirow{2}{*}{Dataset}  & \multicolumn{4}{c|}{Accuracy} \\
 \cline{2-5}
  & SM & LIN & SOP & Ours \\
 \hline
 Oxford-IIIT Pets  & 92.06 & 91.99 & 92.28 & \bf{93.56} \\ 
 Stanford Cars     & 90.78 & 90.83 & 91.04 & \bf{92.60} \\
 \hline
 \end{tabular}
 \vspace{0.2cm}
 \caption{\label{tbl:transfer_learning}{\bf Results on the transfer learning datasets.}}
\end{table}

\subsection{Knowledge Distillation}
We now demonstrate the capabilities of our method in the distillation setting discussed in \S~\ref{sec:distillation}. We used the CIFAR-10 and CIFAR-100 datasets, the softmax CIFAR ResNet-56 models from Tables~\ref{tbl:cifar10_icml}~and~\ref{tbl:cifar100_icml} as the teacher models, and the LeNet-5 network~\cite{Lecun98gradient-basedlearning} as the student model. We use only the cross-entropy loss with the teacher scores with the temperature parameter $T$ set to 20 in all cases. Results are shown in Table~\ref{tbl:distillation}. Once again, significant gains are observed with the kernelized classification layer. This can be attributed to its capabilities to approximate complex teacher probabilities even with weak feature vectors.
\begin{table}[h!]
\centering
 \begin{tabular}{|l | c | c |} 
 \hline
 \multirow{2}{*}{Dataset}  & \multicolumn{2}{c|}{Accuracy} \\
 \cline{2-3}
  & Baseline & Ours \\
 \hline
 CIFAR-10  & 76.06 & \bf{79.85} \\ 
 CIFAR-100 & 44.38 & \bf{46.48} \\
 \hline
 \end{tabular}
 \vspace{0.2cm}
 \caption{\label{tbl:distillation}{\bf Results in the distillation setting.}}
\end{table}

\subsection{Active Learning}
Active learning focuses on reducing human annotation costs by selecting a subset of images to label that are more likely to yield the best model~\cite{settles2009active}. We used different sampling methods such as random, margin~\cite{Lewis1994,Scheffer2001}, and k-center~\cite{sener2017active,Wolf2011} to generate subsets of various sizes. The sampling methods rely on initial seed labels, prediction scores, and distances between the image embeddings; as detailed in the Appendix.

As shown in Table~\ref{tbl:al_cifar100}, our results on random subsets outperform the softmax ResNet56 models on margin and k-center based subsets, and we achieve even better results using improved sampling methods.

\begin{table}[t]
\centering
 \begin{tabular}{|c | c | c | c | c | c | c |} 
 \hline
 \multirow{2}{*}{\%} & \multicolumn{3}{c|}{Baseline}
  & \multicolumn{3}{c|}{Ours} \\
 \cline{2-4}
 \cline{5-7}
 & rnd & mgn & k-ctr & rnd & mgn & k-ctr \\
 \hline
30  & 58.03 & 58.88 & 58.41 & 61.66 & 61.80 & \bf{63.08} \\
40 &  61.05 & 61.81 & 62.02 & 65.25 & 66.28 & \bf{66.35} \\
50 &  64.81 & 65.36 & 65.47 & 67.17 & 68.14 & \bf{69.41} \\
60 &  66.26 & 67.03 & 68.25 & 69.17 & \bf{70.61} & 70.10 \\
70 &  67.47 & 69.16 & 69.84 & 70.06 & 70.90 & \bf{71.50} \\
80 &  69.59 & 69.47 & 71.25 & 71.66 & 72.21 & \bf{72.64} \\
90 &  70.25 & 71.41 & 71.11 & 72.60 & \bf{73.90} & 73.14 \\
 \hline
 \end{tabular}
 \vspace{0.2cm}
 \caption{\label{tbl:al_cifar100}{\bf Active learning on CIFAR-100 dataset.} {\it Terms rnd, mgn, and k-ctr refer to random, margin, and k-center, respectively.}}
\end{table}
\subsection{Ablation Studies}
\label{sec:ablation_study}
For all the ablation experiments below, we used the CIFAR-100 dataset and the ResNet-56 backbone network.
\vspace{-0.2cm}
\subsubsection{Kernel Learning}

We now investigate the benefits of automatic kernel learning compared to using a pre-defined kernel in the kernelized classification layer. To this end, we compare our kernel learning method with two pre-defined kernels in the kernelized classification layer: the polynomial kernel of order $10$ and the Gaussian RBF kernel. 
Results are shown in Table~\ref{tbl:alpha_training}. It is evident that automatically learning the kernel yields substantial improvements. This is not surprising since the search space of the learned kernel contains both Gaussian RBF and polynomial kernels. This shows that optimizing over the all possible positive definite kernels to find the best kernel for the given problem gives significant practical benefits.
\vspace{-0.2cm}
\begin{table}[h!]
\centering
 \begin{tabular}{|l | c |} 
 \hline
 \multicolumn{1}{|c|}{Method}& Accuracy \\
 \hline
 Gaussian RBF kernel &73.21 \\
 Polynomial kernel  & {73.16}  \\ 
 Learned kernel  & {\bf 74.10}  \\ 
 \hline
 \end{tabular}
 \vspace{0.2cm}
 \caption{\label{tbl:alpha_training}{\bf Benefits of learning the best kernel.} \vspace{-0.6cm}}
\end{table}

\subsubsection{Feature rectification}
As discussed previously, different to the usual image classification networks~\cite{resnet}, we remove the $\rm ReLU$ activation from the feature vectors. This is to utilize the full surface of $\Sn$ without restricting ourselves to only the nonnegative orthant. As is evident from Table~\ref{tbl:effect_of_relu}, removing $\rm ReLU$ has only a marginal effect on the standard softmax baseline. It is however an important factor for our method. We consistently used feature vectors without the $\rm ReLU$ activation in all our experiments in the previous sections.
\vspace{-0.2cm}
\begin{table}[h!]
\centering
 \begin{tabular}{|l | c |} 
 \hline
 \multicolumn{1}{|c|}{Method}& Accuracy \\
 \hline
 Softmax classifier with: & \\
 \hspace{0.5cm}rectified features  & 70.96  \\ 
 \hspace{0.5cm}unrectified features  & 71.23  \\ 
 Our classifier with: & \\
 \hspace{0.5cm}rectified features  & 71.61  \\
 \hspace{0.5cm}unrectified features & \bf{74.10} \\
 \hline
 \end{tabular}
 \vspace{0.2cm}
 \caption{\label{tbl:effect_of_relu}{\bf Effect of rectification of the feature vectors.}\vspace{-0.3cm}}
\end{table}

\subsubsection{Activation on the kernel coefficients}
As discussed in \S~\ref{sec:alpha_reg}, we have several choices to impose the non-negative constraint on the kernel coefficients $\alpha_m$s. We experimented with ${\rm ReLU}$, ${\rm sigmoid}$, and ${\rm softmax}$ activations on $\bm{\alpha'}$ and summerized the results in Table~\ref{tbl:alpha_activations}. 

Due to the reasons discussed in \S~\ref{sec:alpha_reg}, we used a weight decay of 0.0001 on the coefficient vector whenever ${\rm ReLU}$ activation is used. Although ${\rm sigmoid}$ and ${\rm softmax}$ activations eliminate the need for weight decay, they put a hard constraint on $|\la \phi(\vec{w}), \phi(\vec{f}) \ra_{\Hi}|$. To overcome this limitation, it is helpful to use a temperature hyperparameter in \eqref{eq:nl_softmax}, where each inner product is divided by $T$ before taking the exponential. We used a temperature of 0.1 and 0.005, with ${\rm sigmoid}$ and ${\rm softmax}$, respectively. Although $\rm{sigmoid}$ gives the best performance in  Table~\ref{tbl:alpha_activations}, we occasionally observed optimization issues with it, which could be due to the vanishing gradient issue associated with this activation function. We therefore stick to ${\rm ReLU}$ in all other experiments. We however note that, in most cases, competitive results can be obtained with ${\rm softmax}$ as well, when used with a temperature of 0.005.

It is also interesting to note that using no activation function on $\bm{\alpha'}$ causes frequent divergence in training. This is consistent with the theory: The summation in \eqref{eq:finite_series} is not guaranteed to be positive definite when $a_m$s are allowed to be negative (see Proposition~\ref{prop:closure}). Therefore, the theory of kernelized classification is not valid in this case.
\begin{table}[h!]
\centering
 \begin{tabular}{|l | c |} 
 \hline
 \multicolumn{1}{|c|}{Activation function}& Accuracy \\
 \hline
 ${\rm sigmoid}$  & {\bf 74.96}  \\ 
 ${\rm softmax}$  & 73.69  \\ 
  ${\rm ReLU}$ & 74.10 \\
  None (linear) & unstable \\
 \hline
 \end{tabular}
 \vspace{0.2cm}
 \caption{{\it \label{tbl:alpha_activations}{\bf Different activation functions on the coefficient vector.} Note that the kernelized classifier is unstable when no activation function is used, this agrees with the theoretical analysis.}\vspace{-0.5cm}}
\end{table}
\section{Conclusion}
We presented a kernelized classification layer for deep neural networks. This classification layer classifies feature vectors in a high dimensional RKHS while automatically learning the optimal kernel that enables this high-dimensional embedding. We showed consistent and substantial accuracy improvements in all experiments: image classification, transfer learning, distillation, and active learning. These accuracy improvements strongly support the usefulness of kernelized classification layer in finding nonlinear patterns in the space of the feature vectors.

{\small
\bibliographystyle{ieee_fullname}
\bibliography{egbib}
}
\appendix

\section*{Appendix A: Synthetic Data Generation}

We generated the blue-orange binary dataset used in \S~\ref{sec:exp:synthetic} of the paper using a mixture of Gaussians. More specifically, we first generated 10 cluster centers for each class by sampling from an isotropic Gaussian distribution with covariance $0.5\,I_3$ and mean $[1, 0, 0]^T$ for the blue class and $[0, 1, 0]^T$ for the orange class. We then generated 5,000 train observations for each class using the following method: for each observation, we uniform-randomly picked a cluster center of the corresponding class and then generated a sample from an isotropic Gaussian distribution centered at that cluster center with covariance $0.02\,I_3$. All the observations were projected on to $S^2$ by $L^2$-normalizing them. The test set was generated in the same manner using the same cluster centers as the train set.

\section*{Appendix B: Experimental Setup}
Throughout the experiments, we use SGD with 0.9 momentum, linear learning rate warmup~\cite{Goyal2017AccurateLM}, cosine learning rate decay~\cite{loshchilov-ICLR17SGDR}, and decide the base learning by cross validation. When a better learning rate schedule is available for the baseline (e.g. the CIFAR schedule in \cite{resnet}), we experimented with both that and our schedule and report the best accuracy of the two. The maximum number of epochs was 450 in all cases. Mini-batch size was 128 for the synthetic and CIFAR datasets and 64 other datasets with larger images. We used the CIFAR data augmentation method in \cite{resnet} for CIFAR-10 and CIFAR-100 datasets, and the Imagenet data augmentation in the same paper for other image datasets.

\section*{Appendix C: Active Learning}
We describe the experimental setting used for active learning here. The goal of this experiment is to show that the kernelized classification layer can produce accurate models even with less data. In order to study this, we produce subsets of datasets under various budgets using several sampling techniques, and evaluate the models trained on this. The simplest one is random sampling, where images are selected randomly under a given budget. Other methods include margin~\cite{Lewis1994,Scheffer2001}, and k-center~\cite{sener2017active,Wolf2011} where class prediction scores and features/embeddings from the images are used in the subset selection. 

We do not rely on the actual labels of the images in the dataset during the subset selection, since active learning is driven toward reducing label annotation costs. We used a 10\% random subset of the original CIFAR-100 dataset with labels to first learn an initial seed model, which was then used to generate predictions and embeddings. Note that only embeddings and class prediction scores from this initial seed model are used in subset selection, and we do not access the original class labels of the images. We use a batch setting where we do not incrementally update the model after selecting every image, and we directly select entire subsets under a given budget. In all our experiments, we used the CIFAR ResNet-56 model. The learning rate, batch size and the number of epochs are provided in Appendix B. The embedding features are of dimension 64. For the k-center method, we need distances between the embeddings, and we used cosine distances computed using the fast similarity search method of~\cite{avq_2020}. 

\end{document}